\documentclass{article}

   \PassOptionsToPackage{numbers, compress}{natbib}

\usepackage[preprint]{neurips_2019}

\usepackage[utf8]{inputenc} 
\usepackage[T1]{fontenc}    
\usepackage{hyperref}       
\usepackage{url}            
\usepackage{booktabs}       
\usepackage{amsfonts}       
\usepackage{nicefrac}       
\usepackage{microtype}      
\usepackage{amsmath}
\usepackage{float}
\usepackage{multirow}
\usepackage{epsfig}
\usepackage{graphicx}
\usepackage{algorithm}
\usepackage{algorithmic}
\usepackage{subfigure}
\usepackage{epstopdf}

\title{Context-Aware Graph Attention Networks}

\author{
  Bo Jiang, Leiling Wang, Jin Tang and Bin Luo\\
  School of Computer Science and Technology, Anhui University\\
  \texttt{jiangbo@ahu.edu.cn} \\
}

\begin{document}

\maketitle

\begin{abstract}
Graph Neural Networks (GNNs) have been widely studied for graph data representation and learning.
However, existing GNNs generally conduct context-aware learning on {node} feature representation only
which usually ignores the learning of {edge} (weight) representation.
In this paper, we propose a novel unified GNN model, named Context-aware Adaptive Graph Attention Network (CaGAT).
CaGAT aims to learn a context-aware attention representation for each graph edge by further exploiting the context relationships among different edges.
In particular, CaGAT conducts \emph{\textbf{context-aware learning}} on both \emph{{node feature representation}} and \emph{{edge (weight) representation}}
 simultaneously and cooperatively in a \emph{unified} manner which can boost their respective performance in network training. 
We apply CaGAT on semi-supervised learning tasks. Promising experimental results on several benchmark datasets demonstrate the effectiveness and benefits of CaGAT.
\end{abstract}

\section{Introduction}


Graph data representation and learning is a fundamental problem in machine learning area.
Recently, Graph Neural Networks (GNNs) have been widely studied for this problem~\cite{defferrard2016convolutional,kipf2016semi,monti2017geometric,hamilton2017inductive,velickovic2017graph,zhou2018graph}.
Overall, existing GNNs can generally be categorized into spatial methods and spectral methods~\cite{zhou2018graph}.

Spectral methods usually define the graph convolution based on spectral representation of graphs.
For example,
Bruna et al.~\cite{bruna2014spectral} propose a spectral based graph convolution network (GCN) by using the eigen-decomposition of graph Laplacian matrix.
Henaff et al.~\cite{henaff2015deep} further introduce a spatially constrained spectral filters in GCN layer-wise propagation.
By using Chebyshev expansion, Defferrard et al.~\cite{defferrard2016convolutional} propose a method to approximate the spectral filters and thus derive a more efficient GCN model. 
By using the first-order approximation of spectral filters, Kipf et al.~\cite{kipf2016semi} also present a more efficient GCN for graph based semi-supervised learning. 

In this paper, we focus on spatial methods. For spatial GNN methods, they generally define the graph convolution  by employing a specific \textbf{node feature diffusion
(or aggregation) operation} on neighbors to obtain the context-aware representations of graph nodes in GNN layer-wise propagation~\cite{monti2017geometric,li2018dcrnn_traffic,CGNN,zhang2018graph,velickovic2018deep}.
For example, Atwood and Towsley~\cite{atwood2016diffusion} propose Diffusion-Convolutional Neural Networks (DCNNs) by
incorporating graph diffusion process into GNN network. 
Li et al.~\cite{li2018dcrnn_traffic} propose Diffusion Convolutional Recurrent Neural Network (DCRNN) which employs a finite K-step truncated random walk
diffusion model for feature aggregation.
Klicpera et al.~\cite{klicpera2018predict} propose to  integrate PageRank propagation into GCN in layer-wise propagation.
Hamilton et al.~\cite{hamilton2017inductive} present  a general inductive representation and learning framework (GraphSAGE) by sampling and
 aggregating features from a node’s local neighborhood.
The above methods generally conduct graph node feature diffusion/aggreagtion on a fixed structure graph.
Recently, Velickovic et al.~\cite{velickovic2017graph} propose Graph Attention Networks (GATs) which first assign different weights for graph edges and
then develop an adaptive weighted feature aggregation mechanism in its layer-wise feature propagation.


\textbf{Motivation}.
One main limitation of the above existing spatial GNNs is that they generally conduct context-aware learning on \textbf{node} feature representation only
which obviously ignores the learning of \textbf{edge} (weight) representation.
Although, the attention learning and representation of edges have been proposed in GATs~\cite{velickovic2017graph}, they are learned independently which
lacks of considering the context relationships among different edges.
 This motivates  us to develop a unified model, named Context-aware adaptive Graph Attention Network (CaGAT),
  which conducts \emph{\textbf{context-aware learning}} on both \emph{\textbf{node feature representation}} and \emph{\textbf{edge (weight) representation}}
  simultaneously in
 layer-wise propagation.

It is known that, in GNNs, the context-aware learning of node feature representation is usually
conducted via a feature diffusion model.
This inspires us to achieve context-aware learning of edge representation similarly by employing a graph diffusion model.
In this paper, 
we adopt a Tensor Product Graph (TPG) diffusion model~\cite{bai2017regularized,yang2012affinity}, which provides
an explicit diffusion for pairwise relationship data.
Specifically, in our CaGAT architecture, we employ TPG and Neighborhood Propagation (NP)~\cite{wang2008label}
 for the context-aware learning of graph edge and node, respectively.
Moreover, both TPG and NP can well be re-formulated as regularization forms,
which further helps us to derive a unified regularization model to integrate them together
for boosting their respective performance.

\textbf{Contributions}. Overall, the main contributions of this paper are summarized as follows:
\begin{itemize}
    \item  We propose a novel spatial GNN, named CaGAT for graph data representation and learning.
  \item  We propose to introduce context-aware edge learning/representation in GNN architecture by employing
  a tensor product graph diffusion model. 
  \item  We provide a unified learning framework to conduct both {edge attention (weight) representation} and {node feature representation}  cooperatively
  in GNN architecture. 
\end{itemize}
Promising experimental results on several benchmark datasets demonstrate the effectiveness of the proposed CaGAT model on
semi-supervised learning tasks. 

\section{Related Works}

Graph Attention Networks (GATs)~\cite{velickovic2017graph} have been widely used for graph data analysis and learning.
GATs conduct two steps in each hidden layer, i.e., 1) graph edge attention estimation  and 2) node feature aggregation and representation.

\textbf{Step 1}: \textbf{Edge attention estimation.}
%
Given a set of node features $\textbf{H}=(\textbf{h}_1,\textbf{h}_2\cdots \textbf{h}_n) \in \mathbb{R}^{d\times n}$ and
graph adjacency matrix $\textbf{A}\in \mathbb{R}^{n\times n}$, GATs define the graph attention $\mathcal{G}(\textbf{h}_i,\textbf{h}_j;\textbf{W},\Theta)$
for each graph edge $e_{ij}$ as,
\begin{align}\label{Eq:gat_def}
\mathcal{G}(\textbf{h}_i,\textbf{h}_j;\textbf{W},\Theta) = \mathrm{softmax}_{G}\big(f(\textbf{W}\textbf{h}_i, \textbf{W}\textbf{h}_j;\Theta)\big)
\end{align}
where  $\textbf{W}\in \mathbb{R}^{\tilde{d}\times d}$ and $\Theta$ denote the layer-specific trainable  parameter of  linear transformation and
 attention estimation, respectively. $\mathrm{softmax}_{G}$ denotes the softmax function defined on graph\footnote{It is  defined as
$$
\mathrm{softmax}_{G}(f_{ij}) = \frac{\exp(f_{ij})}{\sum_{k\in \mathcal{N}_i}\exp(f_{ik})}
$$
where $\mathcal{N}_i$ denotes the neighborhood set of node $v_i$ which is specified by adjacency matrix $\textbf{A}$ of graph.}.
In work~\cite{velickovic2017graph}, the attention mechanism function $f(\textbf{W}\textbf{h}_i, \textbf{W}\textbf{h}_j;\Theta)$
is defined by a single-layer feedforward neural network, parameterized by a weight vector
$\Theta\in \mathbb{R}^{2d\times 1}$.
That is,
\begin{align}\label{Eq:gat_def}
f(\textbf{W}\textbf{h}_i, \textbf{W}\textbf{h}_j;\Theta) = \sigma_a \big(\Theta^{\mathrm{T}} [\textbf{W}\textbf{h}_i\|\textbf{W}\textbf{h}_j]\big)
\end{align}
where $\sigma_a(\cdot)$ denotes some nonlinear function, such as LeakyReLU, and $\|$ denotes the concatenation operation.
%
%

\textbf{Step 2}: \textbf{Note feature aggregation and representation.}
Based on the above graph edge attention,
GATs define the node feature aggregation to obtain the context-aware feature representation  $\textbf{h}'_i$ for
each node $v_i$ as, 
\begin{align}\label{Eq:gat_update}
 \textbf{h}'_i = \sum\nolimits_{j\in\mathcal{N}_i}\mathcal{G}(\textbf{h}_i,\textbf{h}_j;\textbf{W},\Theta)\textbf{W}\textbf{h}_j 
\end{align}
%
In addition, in each hidden layer of GATs, an activation function $\sigma(\cdot)$ is further conducted on $\textbf{h}_i'$ to obtain nonlinear representation.
The last layer of GATs outputs the final representation of graph nodes, which can be used for many learning tasks, such as clustering, visualization and (semi-supervised) classification etc.
In this paper, we focus on semi-supervised classification.
For this task, 
the layer-specific weight parameters  $\{\textbf{W},\Theta\}$ of each layer are optimized by minimizing the cross-entropy loss defined on labelled data, as discussed in works~\cite{kipf2016semi,velickovic2017graph}.


\section{Context-aware Adaptive Graph Attention Network}


There are two main limitations for the above GATs.
First, GATs estimate the attention of of each graph edge independently which ignores the
context relationships among different edges. 
Second, GATs conduct \textbf{Step 1} edge attention estimation and  \textbf{Step 2} node feature aggregation/representation independently which also
neglects the correlation between these two steps and thus may lead to weak suboptimal learning results. 

To overcome these issues, we propose a novel GNN model, named Context-aware adaptive Graph Attention (CaGAT).
CaGAT has two main aspects.
First,  CaGAT aims to learn a context-aware graph edge attention by exploiting the context relationship information of  edges encoded in graph $\textbf{A}$.
Second, CaGAT conducts 1) graph edge attention learning and 2) node feature representation
simultaneously and cooperatively in a \emph{unified} model to boost their respective performance.

\subsection{Context-aware edge attention learning}
%

Setting $\textbf{D}$ as a diagonal matrix with elements $\textbf{D}_{ii}=\sum_j\textbf{A}_{ij}$,
the diffusion matrix can be defined as $\bar{\textbf{A}} =\textbf{D}^{-1}\textbf{A}$.
Motivated by recent works on Tensor Product Graph (TPG) diffusion~\cite{bai2017regularized,yang2012affinity}, we propose to learn a context-aware
graph attention for each edge $e_{ij}$ as follows, 
\begin{align}\label{Eq:cagat_def}
\mathcal{S}^{(t+1)}(\textbf{h}_i,\textbf{h}_j;\textbf{W},\Theta) = \alpha \overbrace{\sum_{h\in \mathcal{N}_i,k\in \mathcal{N}_j}\bar{\textbf{A}}_{ih}\mathcal{S}^{(t)}(\textbf{h}_h,\textbf{h}_k;\textbf{W},\Theta)\bar{\textbf{A}}_{kj}}^{\mathrm{Context\, information}} + (1-\alpha) \overbrace{\mathcal{G}(\textbf{h}_i,\textbf{h}_j;\textbf{W},\Theta)}^{\mathrm{GAT}}
\end{align}
where $t=0,1\cdots T$ and $\mathcal{S}^{(0)}(\textbf{h}_h,\textbf{h}_k;\textbf{W},\Theta) = \mathcal{G}(\textbf{h}_h,\textbf{h}_k;\textbf{W},\Theta)$, as defined in Eq.(1).
The parameter $\alpha\in (0, 1)$ denotes the fraction of attention information that edge $e_{ij}$ receives from its neighbors on graph $\textbf{A}$.
The parameter $\textbf{W}\in \mathbb{R}^{\tilde{d}\times d}$ and $\Theta$ denote the layer-specific trainable  parameter
 of  linear transformation and edge attention estimation, respectively.

\textbf{Remark}. Comparing with GATs, the attention $\mathcal{S}^{(t+1)}(\textbf{h}_i,\textbf{h}_j;\textbf{W},\Theta)$ of edge $e_{ij}$ is
determined based on both its own feature presentation $(\textbf{W}\textbf{h}_i, \textbf{W}\textbf{h}_j)$ and
attentions $\mathcal{G}(\textbf{h}_h,\textbf{h}_k;\textbf{W},\Theta)$ of its neighboring edges $e_{hk}$.
Therefore, CaGAT can capture more context information in graph attention estimation. When $\alpha = 0$, CaGAT degenerates to GATs.
Let $\textbf{S}_{ij}^{(t)} = \mathcal{S}^{(t)}(\textbf{h}_i,\textbf{h}_j;\textbf{W},\Theta)$ and
$\textbf{G}_{ij} = \mathcal{G}(\textbf{h}_i,\textbf{h}_j;\textbf{W},\Theta)$, then Eq.(\ref{Eq:cagat_def}) is compactly formulated as,
\begin{align}\label{Eq:cagat0}
 \textbf{S}^{(t+1)} = \alpha \bar{\textbf{A}}\textbf{S}^{(t)} \bar{\textbf{A}}^{\mathrm{T}}+(1-\alpha)\textbf{G}
 \end{align}
where $t=0,1\cdots T$ and $\textbf{S}^{(0)}  = \textbf{G}$.

\textbf{Regularization framework}.
Here, we show that update Eq.(\ref{Eq:cagat0}) can be theoretically explained from an regularization  framework. 
First, it  can be rewritten as\footnote{For any matrices $\textbf{X},\textbf{Y}$
and $\textbf{Z}$ with appropriate sizes, equation $\mathrm{vec}(\textbf{X}\textbf{Y}\textbf{Z}^{\mathrm{T}})=
(\textbf{Z}\otimes\textbf{X})\mathrm{vec}(\textbf{Y})$ is satisfied. }
\begin{align}\label{Eq:cagat_vec0}
\textrm{vec}(\textbf{S}^{(t+1)}) = \alpha \mathbb{A} \textrm{vec}(\textbf{S}^{(t)}) +(1-\alpha) \textrm{vec}(\textbf{G})
 \end{align}
where $\mathbb{A} = \bar{\textbf{A}}\otimes \bar{\textbf{A}}, \bar{\textbf{A}} =\textbf{D}^{-1}\textbf{A}$, and
$\otimes$ denotes the Kronecker product operation.
The operation $\textrm{vec}(\cdot)$ denotes the column vectorization of an input matrix by stacking its columns one after the next.
Then, one can prove that the converged solution of
Eq.(\ref{Eq:cagat_vec0}) is the optimal solution that minimizes the following
optimization problem~\cite{wang2008label,bai2017regularized}, 
\begin{align}\label{Eq:regcagat0}
\min_{\textbf{S}}\, \mathcal{R}_{\mathrm{CaGAT}}(\textbf{S};\textbf{A},\textbf{G})=  \textrm{vec}(\textbf{S})^{\mathrm{T}}(\textbf{I} - \mathbb{A})\textrm{vec}(\textbf{S})
+\mu\|{\textbf{S}}- {\textbf{G}}\|_F^2
\end{align}
%
where $\mu=\frac{1}{\alpha} - 1$ is a replacement parameter of $\alpha$ to balance two terms.
From this regularization framework,
one can note that, CaGAT aims to learn a context-aware graph attention $\textbf{S}$ by considering the
local consistency as well as preserving the information of original graph attention $\textbf{G}$.


%
%

\subsection{Node feature representation}

%
Based on the proposed context-aware graph attention,
we can obtain the feature representation of each node $v_i$ by using the feature aggregation similar to Eq.(3) as
%
%
\begin{align}\label{Eq:gat_update}
\textbf{h}'_i = \lambda\sum\nolimits_{j\in\mathcal{N}_i}\mathcal{S}(\textbf{h}_i,\textbf{h}_j;\textbf{W},\Theta)\textbf{W}\textbf{h}_j +
(1-\lambda) \textbf{W}\textbf{h}_i
\end{align}
where $\lambda\in (0,1)$ is a weight parameter. 
Using matrix notation, Eq.(8) is formulated as
\begin{align}\label{Eq:gat_update_matrix}
\textbf{H}' = \lambda \textbf{S} \textbf{W}\textbf{H} + (1 - \lambda) \textbf{W}\textbf{H}
\end{align}
where $\textbf{S}_{ij} = \mathcal{S}(\textbf{h}_i,\textbf{h}_j;\textbf{W},\Theta)$ and thus $\sum_{j\in \mathcal{N}_i}\textbf{S}_{ij} = 1, \textbf{S}_{ij}\geq 0$.

\textbf{Regularization framework}. The feature aggregation Eq.(8) can be regarded as a one-step neighborhood propagation (NP) which can also be
derived from the following regularization framework~\cite{wang2008label},
\begin{align}\label{Eq:feature_update_matrix}
\min_{\textbf{H}'} \mathcal{R}_{\mathrm{NP}}(\textbf{H}';\textbf{S},\textbf{W}\textbf{H}) =  \mathrm{Tr}({\textbf{H}'}(\textbf{I} - \textbf{S}){\textbf{H}'}^{\mathrm{T}}) + \gamma\|\textbf{H}' - \textbf{W}\textbf{H}\|^2_F
\end{align}
where $\gamma = \frac{1}{\lambda} - 1$ is a replacement parameter of $\lambda$.

\subsection{Unified model}

Based on regularization frameworks (Eq.(6) and Eq.(9)),
we can propose a \textbf{unified regularization framework} by conducting context-aware learning of both graph edge attention and node feature representation
together as
\begin{align}\label{Eq:unified_model}
\min_{\textbf{S},\textbf{H}'} \ \ \mathcal{U} = \mathcal{R}_{\mathrm{CaGAT}}(\textbf{S};\textbf{A},\textbf{G}) + \beta\mathcal{R}_{\mathrm{NP}}(\textbf{H}';\textbf{S},\textbf{W}\textbf{H})
\end{align}
where $\beta > 0$ balances two terms.
In practical, the optimal $\textbf{S}$ and $\textbf{H}'$ can be obtained via a simple approximate algorithm which alternatively conducts
the following Step 1 and Step 2. 

\textbf{Step 1}. Solving $\textbf{S}$ while fixing $\textbf{H}'$, the problem becomes
\begin{align}\label{Eq:solving_S}
\min_{\textbf{S}} \ \ \mathcal{R}_{\mathrm{CaGAT}}(\textbf{S};\textbf{A},\textbf{G}) + \beta\mathrm{Tr}({\textbf{H}'}(\textbf{I} - \textbf{S}){\textbf{H}'}^{\mathrm{T}})
\end{align}
%
Then, Eq.(12) can be rewritten more compactly as
\begin{align}\label{Eq:solving_S1}
\min_{\textbf{S}}\,\,\, \textrm{vec}(\textbf{S})^{\mathrm{T}}(\textbf{I} - \mathbb{A})\textrm{vec}(\textbf{S})
+\mu\|{\textbf{S}}- {\textbf{G}}\|_F^2- \beta \mathrm{Tr} ( {\textbf{S}}{{\textbf{H}'}^{\mathrm{T}}\textbf{H}'})
\end{align}
%
It is equivalent to
\begin{align}\label{Eq:solving_S11}
\min_{\textbf{S}}\,\,\, \textrm{vec}(\textbf{S})^{\mathrm{T}}(\textbf{I} - \mathbb{A})\textrm{vec}(\textbf{S})
+\mu\|{\textbf{S}}- ({\textbf{G}}+ \textstyle \frac{\beta}{2\mu} {{\textbf{H}'}^{\mathrm{T}}\textbf{H}'})\|_F^2 
\end{align}
%
The optimal $\textbf{S}$ can be computed approximately via a power iteration update algorithm as
\begin{align}\label{Eq:cagat000}
 \textbf{S}^{(t+1)} = \alpha \bar{\textbf{A}}\textbf{S}^{(t)} \bar{\textbf{A}}^{\mathrm{T}}+(1-\alpha)\textbf{G} + \xi {\textbf{H}'}^{\mathrm{T}}\textbf{H}'
 \end{align}
where $t=0,1\cdots T$ and $\textbf{S}^{(0)}  = \textbf{G}$. Parameter $\alpha=\frac{1}{1+\mu}$ and $\xi =\frac{(1-\alpha)\beta}{2\mu}$.

\textbf{Step 2}. Solving $\textbf{H}'$ while fixing $\textbf{S}$, the problem becomes Eq.(10).
The optimal solution is
\begin{align}\label{Eq:solving_H0}
\textbf{H}' = (1-\lambda)(\textbf{I} - \lambda \textbf{S})^{-1}\textbf{WH}
\end{align}
and an approximate solution can be obtained by~\cite{wang2008label,zhou2004learning}
\begin{align}\label{Eq:solving_H}
\textbf{H}' = \big[(\lambda\textbf{S})^T +(1-\lambda)\textstyle\sum^{T-1}_{t=0}(\lambda\textbf{S})^i\big] \textbf{WH}
\end{align}
When $T=1$, we obtain the one-step iteration solution as
\begin{align}
\textbf{H}' = \lambda \textbf{S} \textbf{W}\textbf{H} + (1 - \lambda) \textbf{W}\textbf{H}
\end{align}
%

\subsection{CaGAT architecture}

The overall layer-wise propagation of the proposed CaGAT is summarized in Algorithm 1, where $\sigma(\cdot)$ used in the last step denotes an activation function, such as $\mathrm{ReLU}(\cdot) = \max(0,\cdot)$.
Considering the efficiency of CaGAT training, we employ a truncated iteration algorithm to optimize the context-aware problem approximately in CaGAT architecture.
In this paper, we apply CaGAT on semi-supervised classification.
Similar to many other works~\cite{kipf2016semi,velickovic2017graph}, the optimal weight matrix $\textbf{W}$ and weight vector $\Theta$ of  each hidden layer in CaGAT 
are trained by minimizing the cross-entropy loss via an Adam algorithm~\cite{Adam} which is initialized by using Glorot initialization~\cite{glorot2010understanding}.
Figure 1 shows the training loss values across different epochs.
One can note that, CaGAT obtains obviously lower cross-entropy loss values than GAT at convergence, which clearly demonstrates the higher predictive accuracy of CaGAT model.

\begin{algorithm}[h]
\caption{CaGAT layer-wise propagation}
\begin{algorithmic}[1]
\STATE \textbf{Input:} Feature matrix $\textbf{H}\in \mathbb{R}^{d\times n}$, graph ${\textbf{A}}\in \mathbb{R}^{n\times n}$ and weight vector $\Theta$, network weight matrix $\textbf{W}$.   parameter $\xi,\alpha$ and $\lambda$,  maximum iteration $K$ and $T$
\STATE \textbf{Output:} Feature map $\textbf{H}'$ 
\STATE Compute diffusion matrix $\bar{\textbf{A}} = \textbf{D}^{-1}\textbf{A}$
\STATE Compute graph attention $\textbf{G}_{ij}$ as
$
\textbf{G}_{ij} = \mathcal{G}(\textbf{h}_i,\textbf{h}_j;\textbf{W},\Theta)
$ (Eq.(1))
\STATE Update ${\textbf{H}'}$ as
$
\textbf{H}'  \leftarrow \lambda \textbf{G} \textbf{W}\textbf{H} + (1 - \lambda) \textbf{W}\textbf{H}
$
\STATE  Initialize $\textbf{S}$ = $ \textbf{G}$
\FOR {$k=1,2\cdots K$}
\STATE  \emph{Compute CaGAT} 
\FOR {$t=1,2\cdots T$}
\STATE $\textbf{S} \leftarrow \alpha \bar{\textbf{A}}\textbf{S}\bar{\textbf{A}}^{\mathrm{T}}+(1-\alpha)\textbf{G} + \xi {\textbf{H}'}^{\mathrm{T}}\textbf{H}'
$
\ENDFOR  \label{code:recentEnd}
\STATE  \emph{Compute feature aggregation}
\STATE 
$
\textbf{H}'  \leftarrow \lambda \textbf{S} \textbf{W}\textbf{H}' + (1 - \lambda) \textbf{W}\textbf{H}
$
\ENDFOR \label{code:recentEnd}
\STATE Return $\textbf{H}' \leftarrow \sigma(\textbf{H}')$
\emph{}
\end{algorithmic}
\end{algorithm}
%
\begin{figure}[htpb]
\centering
\centering
\includegraphics[width=0.5\textwidth]{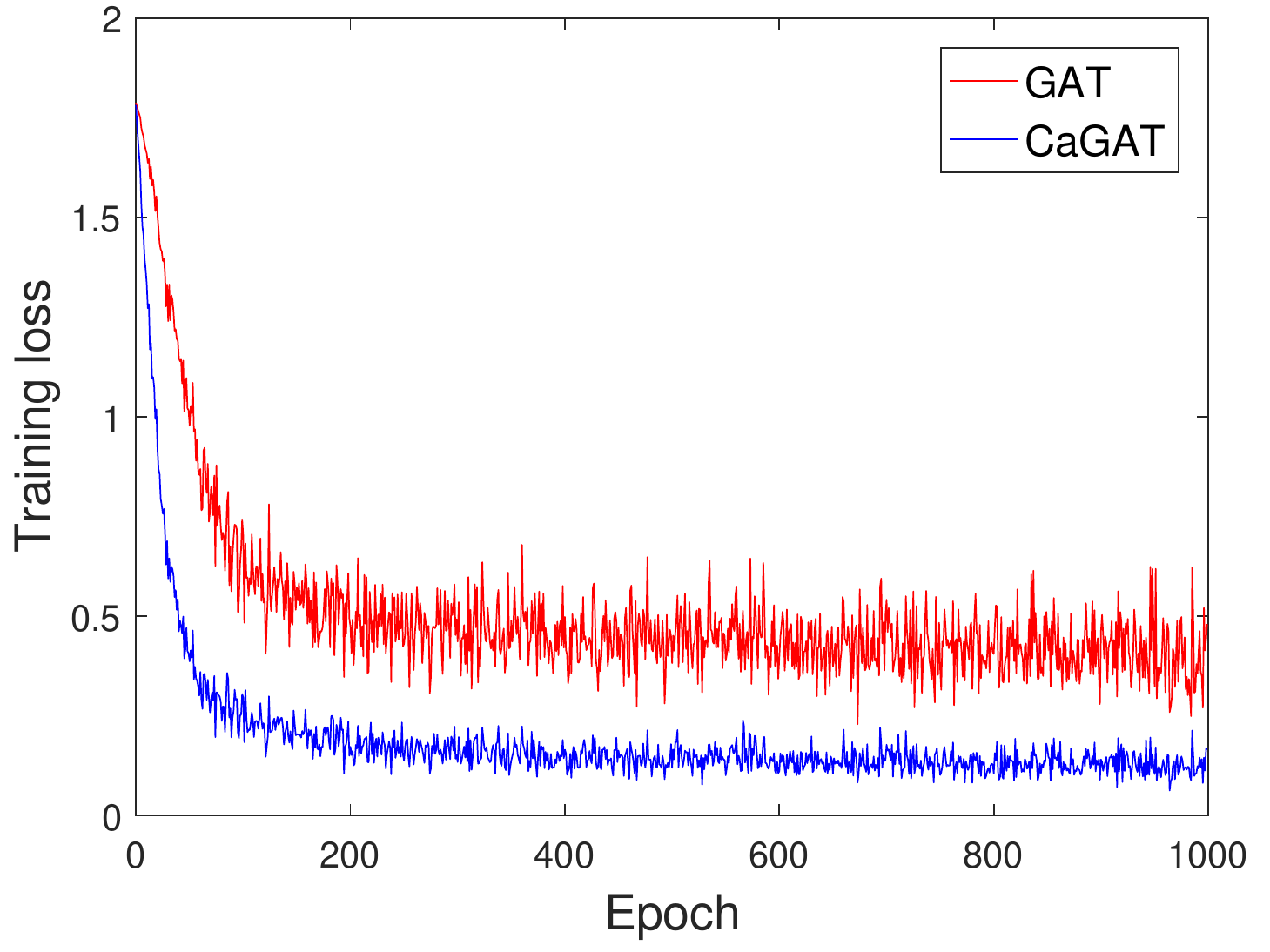}
  \caption{Demonstration of cross-entropy loss values across different epochs on Cora dataset.}\label{fig::lambda}
\end{figure}
\textbf{Complexity analysis}.
The main computation complexity of CaGAT network training involves context-aware attention computation and feature aggregation.
For context-aware attention computation step, we adopt a simple update algorithm and the whole computational complexity is $\mathcal{O}(Tn^3)$ in the worst case (for the full dense graph).
For feature aggregation step, we adopt a similar feature aggregation used in GATs~\cite{velickovic2017graph}, which has the computational complexity as $\mathcal{O}(n^2d)$ in the worst case, where $d$ denotes the feature dimension of $\textbf{H}$.
Therefore, the whole complexity is $\mathcal{O}(K(Tn^3+n^2d))$.
In our experiments, the maximum iterations $\{K, T\}$ are set to $\{3, 2\}$ respectively. Thus, the overall layer-wise propagation in CaGAT is not computationally expensive.

\section{Experiments}

To verify the effectiveness of the proposed method, we implement our CaGAT and test it on four benchmark datasets and compare it with some other related methods.

\subsection{Datasets}

We test our method on four benchmark datasets, including Cora, Citeseer, Pubmed from a graph neural network dataset~\cite{shchur2018pitfalls} and Cora-ML~\cite{mccallum2000automating,klicpera_predict_2019}.
The details of these datasets and their usages in our experiments are introduced below. 

\noindent\textbf{Cora.}
This data contains 2485 nodes and 7554 edges. Nodes correspond to documents and edges to citations between documents.
Each node has a 1433 dimension feature descriptor and all the nodes are classified into 7 classes.

\noindent\textbf{Citeseer.}
This data contains 2110 nodes and 5778 edges. Each node corresponds to a document and edge to citation relationship between documents.
The nodes of this network are classified into 6 classes and each node has been represented by a 3703 dimension feature descriptor.

\noindent\textbf{Pubmed.}
This dataset contains 19717 nodes and 64041 edges which are classified into 3 classes.
Each node is represented by a 500 dimension feature descriptor.

\noindent\textbf{Cora-ML.}
It contains 2810 nodes and 7981 edges.
Each node is represented by a 2879 dimension feature descriptor and all the nodes are falling into 7 classes.

\begin{table*}[!htp]
\centering
\caption{Comparison results of different methods on four benchmark datasets. The best results are marked as bold. }
\centering
\begin{tabular}{c||c|c||c|c}
  \hline
  \hline
  Dataset& \multicolumn{ 2}{c||}{Cora} & \multicolumn{ 2}{c}{Citeseer} \\
  \hline
  No. of label (each class) & 10 & 20  & 10 & 20  \\
  \hline
  MLP  & 50.30 $\pm$ 2.98  & 58.34 $\pm$ 1.84 & 52.31$\pm$ 2.52 & 58.98 $\pm$ 1.89 \\
  LogReg~\cite{shchur2018pitfalls}  & 50.78 $\pm$ 4.66 & 58.38 $\pm$ 2.36 & 53.23 $\pm$ 3.28 & 60.86 $\pm$ 2.77 \\
  LabelProp~\cite{zhu2003semi}  & 67.78 $\pm$ 4.62 & 75.41 $\pm$ 2.75 & 65.42 $\pm$ 2.65 & 68.24 $\pm$ 2.07 \\
  LabelProp NL~\cite{zhu2003semi}  & 70.59 $\pm$ 1.56 & 74.36 $\pm$ 1.69 & 63.81$\pm$ 2.48 & 66.61 $\pm$ 1.82 \\
  DGI~\cite{velickovic2018deep}  & 70.78 $\pm$ 4.46 & 71.19 $\pm$ 4.55 & 68.69 $\pm$ 2.35 & 67.97 $\pm$ 2.66 \\
  GraphSAGE~\cite{hamilton2017inductive}  & 70.89 $\pm$ 3.99 & 77.54 $\pm$ 1.91 & 68.48 $\pm$ 1.92 & 71.84 $\pm$ 1.36 \\
  CVD+PP~\cite{chen2017stochastic} & 75.97 $\pm$ 2.65 & 79.57 $\pm$ 1.18 & 70.74 $\pm$ 1.74 & 70.95 $\pm$ 1.21 \\
  GCN~\cite{kipf2016semi} & 76.35 $\pm$ 2.79 & 79.03 $\pm$ 1.52 & 71.04 $\pm$ 1.79  & 71.49 $\pm$ 1.24 \\
  GATs~\cite{velickovic2017graph} & 76.54 $\pm$ 2.50  & 79.07 $\pm$ 1.36 & 71.07 $\pm$ 1.23 & 71.91 $\pm$ 1.27  \\
  \hline
  CaGAT & \textbf{ 77.98 $\pm$ 2.38 } & \textbf{ 80.51 $\pm$ 0.80 }  & \textbf{ 72.12 $\pm$ 1.35 } & \textbf{ 73.21 $\pm$ 1.36 }  \\
  \hline
  \hline
  Dataset & \multicolumn{ 2}{c||}{Pubmed} & \multicolumn{ 2}{c}{Cora-ML}\\
  \hline
  No. of label (each class)  & 10 & 20 & 10 & 20  \\
  \hline
  MLP  & 64.39 $\pm$ 2.77 & 69.76 $\pm$ 1.81 & 53.87 $\pm$ 2.54 & 63.76 $\pm$ 1.75 \\
  LogReg~\cite{shchur2018pitfalls}  & 60.75 $\pm$ 3.59 & 64.03 $\pm$ 2.89  & 48.07 $\pm$ 6.34 & 62.17 $\pm$ 2.64 \\
  LabelProp~\cite{zhu2003semi}  & 64.03 $\pm$ 8.72 & 70.64 $\pm$ 5.20  & 65.03 $\pm$ 4.41 & 71.81 $\pm$ 3.78 \\
  LabelProp NL~\cite{zhu2003semi}  & 68.01 $\pm$ 5.03 & 73.31 $\pm$ 1.47 & 72.23 $\pm$ 2.18 & 75.26 $\pm$ 1.48  \\
  DGI~\cite{velickovic2018deep}  & 70.08 $\pm$ 2.71 & 71.62 $\pm$ 1.59 & 69.38 $\pm$ 4.99 & 72.26 $\pm$ 2.89 \\
  GraphSAGE~\cite{hamilton2017inductive}  & 69.92 $\pm$ 3.85 & 73.16 $\pm$ 2.08 & 75.24 $\pm$ 3.65 & 80.82 $\pm$ 1.75 \\
  CVD+PP~\cite{chen2017stochastic} & 75.00 $\pm$ 3.17 & 76.89 $\pm$ 1.70 & 78.51 $\pm$ 2.64 & 80.83 $\pm$ 1.54 \\
  GCN~\cite{kipf2016semi} & 74.88 $\pm$ 2.85 & 77.41 $\pm$ 2.06 & 78.41 $\pm$ 2.37  &  80.04 $\pm$ 1.63  \\
  GATs~\cite{velickovic2017graph} & 74.64 $\pm$ 2.23 & 76.74 $\pm$ 1.73 & 78.98 $\pm$ 2.09 & 80.33 $\pm$ 1.98 \\
  \hline
  CaGAT & \textbf{ 75.63 $\pm$ 3.09 } & \textbf{ 77.68 $\pm$ 2.15}  & \textbf{ 80.38 $\pm$ 2.09 } & \textbf{ 81.86 $\pm$ 1.85 }  \\
  \hline
  \hline
\end{tabular}
\end{table*}

\subsection{Experimental setting}


For all datasets, we randomly select 10 and 20 samples in each class as labeled  data  for training the network
and use the other 20 labeled data in each class for validation purpose.
 The remaining unlabeled samples are used as testing samples.
All the reported results are averaged over ten runs  with different groups of training, validation and testing data splits.

Similar to traditional GATs~\cite{velickovic2017graph}, the number of hidden convolution layers in CaGAT is set as 2.
The number of units in each hidden layer is set as 8 and it also has eight head-attentions, as suggested in GATs~\cite{velickovic2017graph}.
We train our CaGAT for a maximum of 10000 epochs (training
iterations) by using an ADAM algorithm~\cite{Adam} with a learning rate of 0.005.
We stop training if the validation loss does not decrease for 100 consecutive epochs,
as suggested in work~\cite{velickovic2017graph}.
All the network weights $\{\textbf{W},\Theta\}$ of each hidden layer are initialized by using Glorot initialization~\cite{glorot2010understanding}.
The balanced parameter $\alpha$ and $\xi$ (Eq.(15)) in CaGAT are set to 0.4 and 0.001, respectively.   
The parameter $\lambda$ (Eq.(18)) is set to 0.3.
 We will shown in \S 4.4 that CaGAT is generally insensitive w.r.t. these parameters.

\subsection{Comparison results}

\textbf{Baselines.}  We first compare our CaGAT model with the baseline model GATs~\cite{velickovic2017graph} which is the most related model with our CaGAT.
We also compare our method against some other  related  graph approaches which contain
i)  Graph based semi-supervised learning method Label Propagation(LabelProp)~\cite{zhu2003semi} and Normalized Laplacian Label Propagation (LabelProp NL)~\cite{zhu2003semi},
 ii)  Attribute-based models like Logistic Regression (LogReg) and Multilayer Perceptron (MLP)~\cite{shchur2018pitfalls} that do not consider the graph structure
and iii)  Graph neural network methods including Graph Convolutional Network (GCN)~\cite{kipf2016semi}, Graph Attention Networks (GATs)~\cite{velickovic2017graph}, Deep Graph Informax(DGI)~\cite{velickovic2018deep}, GraphSAGE~\cite{hamilton2017inductive} and CVD+PP~\cite{chen2017stochastic}
The codes of these comparison methods are available and we use them in our experiments.


%
\textbf{Comparison results.} Table 1 summarizes the comparison results on four benchmark datasets. 
The best results are marked as bold.
%
Here, we can note that,
(1) CaGAT consistently outperforms the baseline method GATs~\cite{velickovic2017graph} on all datasets.
It clearly demonstrates the effectiveness and benefits of the proposed context-aware graph attention estimation on conducting graph data learning. 
%
(2) CaGAT outperforms recent graph neural network method GraphSAGE~\cite{hamilton2017inductive}, APPNP~\cite{klicpera2018predict} and DGI~\cite{velickovic2018deep},
which demonstrates the advantages of CaGAT on graph data representation and semi-supervised learning.
(3) CaGAT can obtain better performance than other semi-supervised learning methods, such as
LabelProp~\cite{zhu2003semi}, Normalized Laplacian Label Propagation (LabelProp NL)~\cite{zhu2003semi}, LogReg and Multilayer Perceptron (MLP)~\cite{shchur2018pitfalls}.
 It further demonstrates the effectiveness of CaGAT on conducting semi-supervised learning tasks.

\begin{figure}[!htp]
\centering
\centering
\subfigure[Results across different parameter $\alpha$ values]{\includegraphics[width=0.48\textwidth]{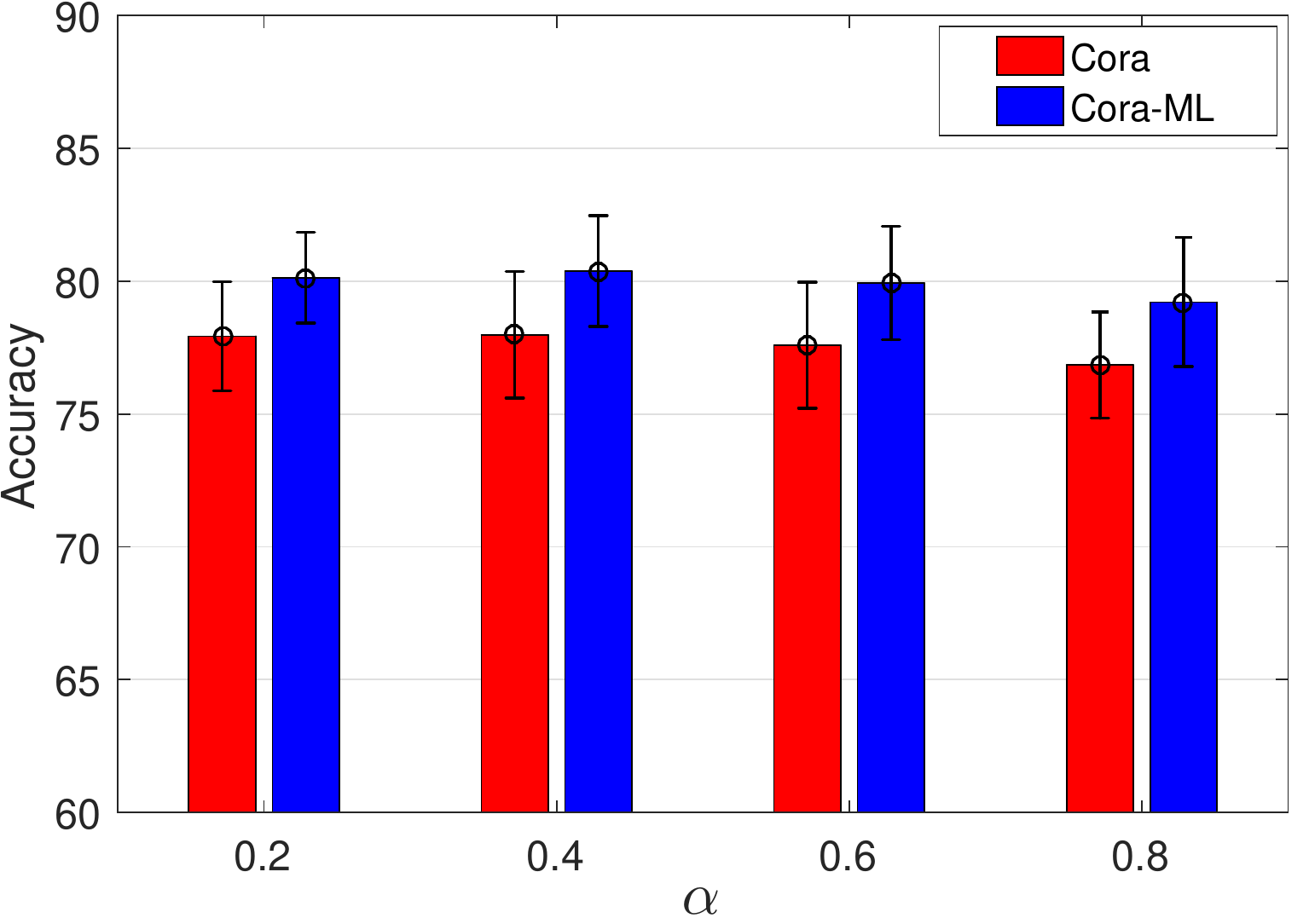}}\subfigure[Results across different parameter $\lambda$ values]{\includegraphics[width=0.48\textwidth]{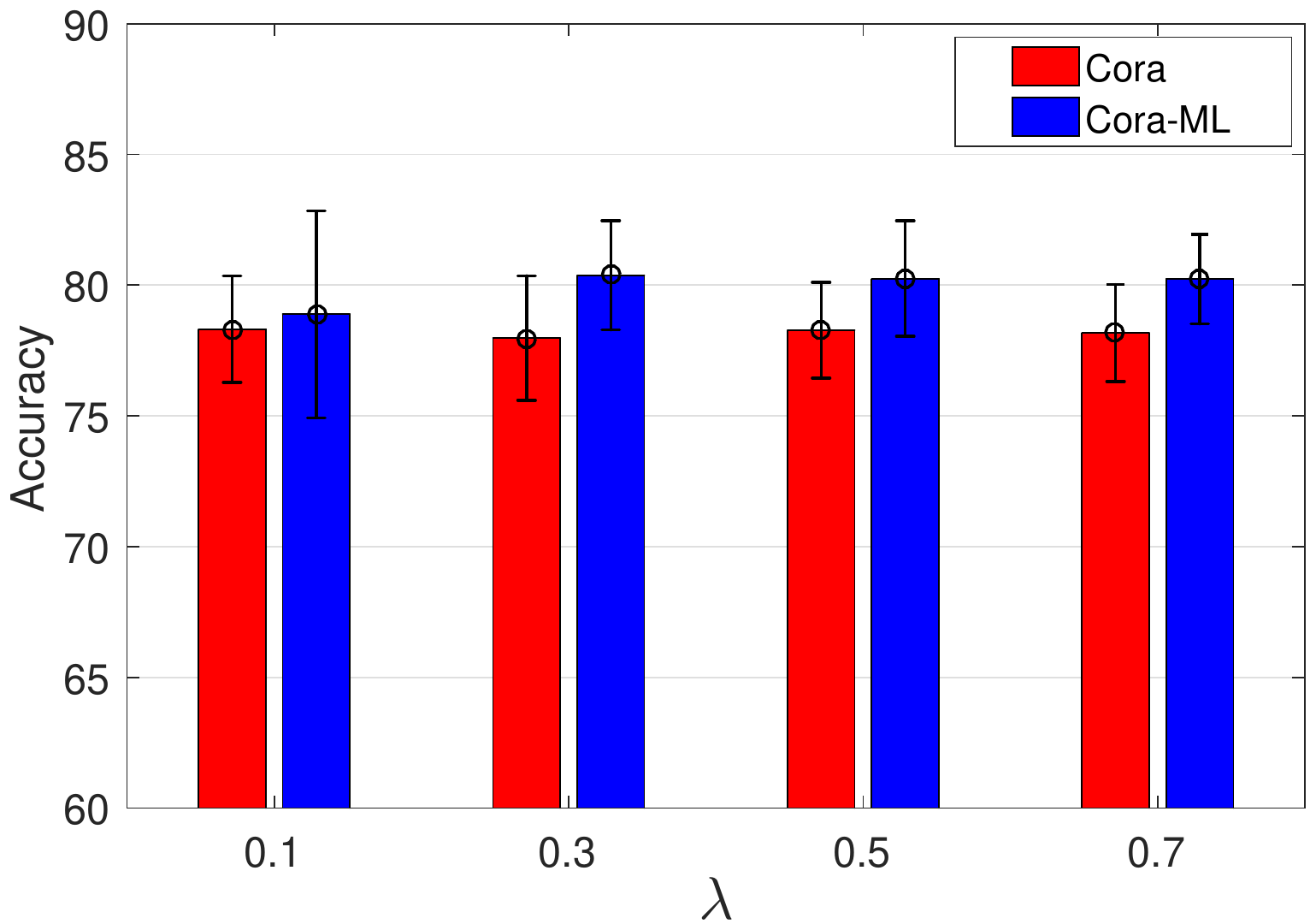}}
  \caption{Results of CaGAT with different settings of parameter $\alpha$ and $\lambda$.}\label{fig::lambda}
\end{figure}
\begin{table}[!htp]
\centering
\caption{Results of CaGAT across different parameter $\xi$ values. }\label{tb::patch}
\centering
\begin{tabular}{c|c|c|c|c|c}
\hline
\hline
  $\xi$ & 1e-2 & 1e-3 & 1e-4 & 1e-5 & 0  \\
 \hline
  Cora & 77.49 $\pm$ 2.35 & 77.98 $\pm$ 2.38 & 77.14 $\pm$ 2.44 & 77.02 $\pm$ 2.48 & 77.02 $\pm$ 2.48 \\
\hline
  Cora-ML  & 79.43 $\pm$ 2.02 & 80.38 $\pm$ 2.09 & 77.21 $\pm$ 2.36 & 77.03 $\pm$ 2.38 & 77.05 $\pm$ 2.37 \\
\hline
\hline
\end{tabular}
\end{table}
\subsection{Parameter analysis}

There are three main parameters $\{\lambda,\alpha, \xi\}$ of the proposed CaGAT in which $\alpha$ is used to weight the importance of graph attention in context-aware attention estimation
 while $\xi$ is used to balance the graph attention and feature aggregation terms in the proposed unified model.
%
%
%
Table 2 shows the performance of CaGAT model under different  parameter $\xi$ values (a replacement of parameter $\beta$ in Eq.(11)).
Note that, when $\xi=0$, the proposed model degenerates to conduct graph attention and feature aggregation independently in network training.
From Table 2, we can note that
(1) CaGAT with $\xi>0$ outperforms that with  $\xi=0$, which clearly demonstrates the desired benefits of the
proposed unified cooperative learning manner to boost the performance of both edge and node learning.
(2) CaGAT is generally insensitive w.r.t. parameter $\xi$. It can obtain better results as $\xi$ varying in parameter range 1e-3$\sim$1e-2.
Figure 2 shows the performance of CaGAT model under different parameter $\alpha$ and $\lambda$ values, respectively.
Here, we can note that, CaGAT is generally insensitive w.r.t. parameter $\alpha$ and $\lambda$, and  obtains
 better results as $\alpha$ varying in range $0.2\sim0.6$ and $\lambda$ varying in range $0.1\sim0.7$.


\section{Conclusion}

In this paper, we propose a novel spatial graph neural network, named Context-aware Adaptive Graph Attention Network (CaGAT).
The key idea behind CaGAT is to compute a context-aware graph attention  by employing a Tensor Product Graph (TPG) diffusion technique.
Moreover, CaGAT conducts graph edge attention learning and node feature representation
cooperatively in a \emph{unified} scheme which can further exploit the correlation between them in
 network training and thus can boost their respective performance.
Experimental results on four widely used benchmarks validates the benefits of CaGAT on semi-supervised learning tasks.


\bibliographystyle{ieee}
\bibliography{nmfgm-CaGAT}

\end{document}